\documentclass[11pt]{article}
\pdfoutput=1  % arXiv: force pdfLaTeX (PDF/PNG figures)

\usepackage[preprint]{acl}      % arXiv preprint version (de-anonymized)

\usepackage{times}
\usepackage{latexsym}
\usepackage[T1]{fontenc}
\usepackage[utf8]{inputenc}
\usepackage{microtype}
\usepackage{inconsolata}

\usepackage{booktabs}
\usepackage{array}
\usepackage{amsmath}
\usepackage{amssymb}
\usepackage{amsthm}
\usepackage{graphicx}
\usepackage{float}
\usepackage{xcolor}
\usepackage{tikz}
\usepackage{tabularx}
\usepackage{algorithm}
\usepackage{algpseudocode}

\usetikzlibrary{positioning,fit,arrows.meta,calc,shapes.geometric,backgrounds}

\usepackage{hyperref}

\hypersetup{
    colorlinks=true,
    linkcolor=blue,
    citecolor=blue,
    urlcolor=blue
}

\title{Semantic Consistency Policy Optimization for Reinforcement Learning of LLM Agents}

\author{
  Peng Xu$^{\dagger}$ \quad Sijia Chen$^{\dagger}$ \quad Junzhuo Li$^{\dagger\ddagger}$ \quad Xuming Hu$^{\dagger\ddagger}$ \\
  $^{\dagger}$The Hong Kong University of Science and Technology (Guangzhou) \\
  $^{\ddagger}$The Hong Kong University of Science and Technology \\
  \texttt{\{pxu364, jz.li\}@connect.hkust-gz.edu.cn} \\
  \texttt{\{sijiachen, xuminghu\}@hkust-gz.edu.cn}
}

\begin{document}
\maketitle

% =============================================================
% =============================================================
\begin{abstract}

Group-based reinforcement learning effectively post-trains LLM agents for long-horizon, sparse-reward tasks by deriving step-level credit from trajectory outcomes. 
However, this ties a step's credit to its rollout's final outcome: semantically near-identical intermediate steps receive opposite credit depending on whether their trajectory eventually succeeded or failed. 
Such semantic credit inconsistency sends conflicting gradients to similar actions and wastes the partially-correct progress inside failed rollouts. Motivated by this, we propose \textbf{S}emantic \textbf{C}onsistency \textbf{P}olicy \textbf{O}ptimization (\textbf{SCPO}), a value-free reward-shaping method that mitigates this inconsistency by recovering step-level credit from successful siblings in the same rollout group.
Concretely, SCPO scores each failed step against a successful sibling and adds positive step-level credit for new progress along that sibling.
On ALFWorld and WebShop, SCPO matches or exceeds strong group-based baselines, reaching \textbf{93.7$\pm$4.1\%} success on ALFWorld and \textbf{74.8$\pm$2.0\%} on WebShop at 1.5B parameters, with gains concentrated on the hardest multi-step tasks.

\end{abstract}

\section{Introduction}
\label{sec:intro}

Large language models (LLMs) increasingly act as \emph{agents} that perceive, reason, and act in open-ended, interactive environments~\citep{achiam2023gpt4,geminiteam2024gemini}.
Representative applications include embodied assistants in simulated households~\citep{shridhar2021alfworld}, web navigators that complete multi-step browsing tasks~\citep{webshop}, and tool-using agents for search and reasoning~\citep{toolrl,searchr1}.
Beyond single-turn language understanding, such agents must perform long-horizon planning and robust decision-making, repeatedly observing the environment, acting, and recovering from earlier mistakes, often before any reward is observed.

Reinforcement learning (RL) has become a key paradigm for post-training LLMs~\citep{ouyang2022instructgpt,deepseekr1}.
In particular, group-based, value-free methods such as RLOO~\citep{kool2019rloo}, GRPO~\citep{deepseekmath}, and DAPO~\citep{dapo2025} estimate relative advantages within a group of sampled rollouts instead of training a separate critic, achieving strong performance at low computational cost.
These methods are most established on single-turn tasks such as mathematical reasoning and code generation, where reward is immediate and credit assignment~\citep{pignatelli2024tca} is straightforward.
In multi-turn agentic settings, however, the reward is typically a single terminal signal---often a binary success/failure delivered only after a long trajectory of tens of steps---so applying these algorithms naively assigns every step the same trajectory-level advantage and collapses step-level distinctions.
To recover fine-grained credit without a critic or additional rollouts, GiGPO~\citep{feng2025gigpo} groups steps that revisit the same environment state across rollouts and estimates step-level relative advantages within these state-aligned groups, and subsequent work has further enriched group-based agentic RL (e.g., HGPO~\citep{he2026hgpo}, HCAPO~\citep{tan2026hcapo}, and Context-Lite~\citep{contextlite2025}).

However, these step-level remedies still derive a step's credit from its trajectory's eventual outcome: a locally correct step is penalised whenever its rollout fails at a later step, even when it shares a stage with successful rollouts. 
Two semantically near-identical steps---one from a successful rollout, one from a rollout that erred only later---therefore receive opposite training signals, decided by their trajectories' outcomes rather than by the step itself.
We call this failure mode \emph{semantic credit inconsistency}: it sends conflicting gradients to similar actions and wastes the partially correct progress hidden inside failed rollouts.

\begin{figure*}[t]
\centering
\includegraphics[width=\textwidth]{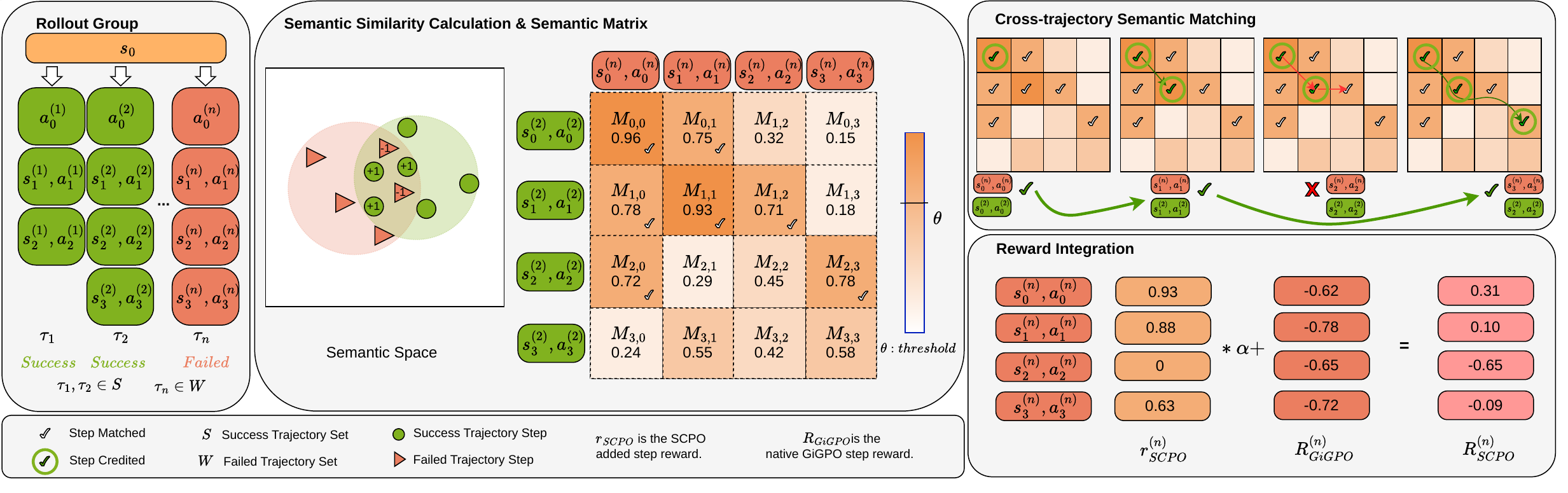}
\caption{\textbf{Overview of SCPO}, a reward-level plugin on group-based agentic RL, read left to right. \textbf{(1)~Rollout group:} for each task we sample a group of trajectories and take a successful sibling as the \emph{success reference}. \textbf{(2)~Semantic similarity:} a frozen cross-encoder scores every failed step against every reference step, forming a similarity matrix $M$. \textbf{(3)~Cross-trajectory matching:} a failed step is \emph{matched} to a reference step when their similarity exceeds the threshold $\theta$, and \emph{credited} only if that position has not already been credited---so credit advances monotonically through the reference and no position is rewarded twice. \textbf{(4)~Reward integration:} the recovered credit, scaled by $\alpha$, is added to the native GiGPO step reward and passed to the advantage estimator.}
\label{fig:pipeline}
\end{figure*}

To address this issue, we propose \textbf{S}emantic \textbf{C}onsistency \textbf{P}olicy \textbf{O}ptimization (\textbf{SCPO}), a reward-level plugin inserted between step-reward construction and advantage estimation, which can in principle also be combined with any group-based agentic RL method; Figure~\ref{fig:pipeline} gives an overview.
Concretely, SCPO treats an in-group successful sibling as a \emph{reference} and rewards a failed step for the new semantic progress it makes toward that reference, crediting each reference position at most once.

We evaluate SCPO on ALFWorld and WebShop with Qwen2.5-1.5B-Instruct and Qwen2.5-7B-Instruct.
The gains are largest at 1.5B---$+7.0$ on ALFWorld and $+9.8$ on WebShop task-success over the GiGPO base it wraps, concentrated in the harder multi-step families; at 7B the margin narrows but SCPO stays competitive with the strongest published baselines.
Our contributions are threefold:
(i) we identify \emph{semantic credit inconsistency} as a credit-assignment failure mode in sparse-reward agentic RL;
(ii) we propose SCPO, a value-free reward shaper that recovers step-level credit through monotonic semantic matching against successful in-group siblings;
and (iii) to our knowledge, we are the first to bring frozen cross-encoder semantic step matching into agentic RL, identifying useful behaviour in failed trajectories without training an additional verifier or reward model.
Our code, training logs, and run records will be released after submission.

\section{Related Work}
\label{sec:related}

\paragraph{Group-based RL for LLM post-training.} Group-based, value-free RL methods compute relative advantages within rollout groups instead of training a critic. GRPO~\citep{deepseekmath} introduced trajectory-level relative advantages over multiple sampled responses, and RLOO~\citep{kool2019rloo} and DAPO~\citep{dapo2025} have been widely used for reasoning and code post-training. 
In multi-turn agentic settings, GiGPO~\citep{feng2025gigpo} adds step-level advantages by grouping steps that revisit the same \emph{anchor} state across rollouts, while HGPO~\citep{he2026hgpo} and Context-Lite~\citep{contextlite2025} refine context- and token-level advantage computation and EMPG~\citep{empg2025} modulates the step-level gradient by stepwise uncertainty. 
SCPO intervenes at the step-reward level, recovering credit for failed steps from successful in-group siblings.

\paragraph{Process supervision and hindsight credit assignment.} A direct remedy for sparse outcome reward and the temporal credit-assignment problem~\citep{pignatelli2024tca} is to add step-level supervision. 
Process reward methods provide dense signals through learned critics, rollout statistics, demonstrations, preference objectives, or LLM/verifier calls~\citep{choudhury2024agentprm,fudan2025agentprm,istar2025oprl,xia2025agentrm,rtmc2025,zhang2025agentvrl,spa_rl2025}, with recent agentic variants spanning programmatic meta-reasoning rewards~\citep{rlvmr2025}, step-wise search rewards~\citep{stepsearch2025}, and segment-level preference optimization~\citep{sdpo2025}; these gains, however, come at the cost of training a process reward model, collecting demonstrations, or expensive inference-time verification. 
Hindsight credit assignment instead reinterprets earlier actions after observing the outcome: classical HCA~\citep{harutyunyan2019hca} estimates hindsight-conditioned action likelihoods, and HCAPO~\citep{tan2026hcapo} uses the policy LLM as a post-hoc critic---trading the reward model for causal-probability estimation or LLM judging. 
SCPO follows the same hindsight spirit by treating a successful in-group sibling as the hindsight reference against which failed steps are credited, without training a task-specific process reward model or invoking an LLM critic.

\paragraph{Reward shaping and successful-behaviour reuse.} A related line reshapes rewards or reuses good behaviour. 
Potential-based reward shaping preserves optimal policies under a state-potential form~\citep{ng1999pbrs}, and recent work shows reward shaping is an efficient lever for agentic RL, especially for smaller policies~\citep{plannerr1}. 
Self-imitation learning~\citep{oh2018sil} and imitation methods such as GAIL~\citep{ho2016gail} reuse high-return or expert behaviour, and agentic variants such as SPEAR~\citep{spear2025} replay whole successful trajectories from a buffer. 
A separate strand turns trajectory or text similarity into a learning signal, e.g.\ BERTScore~\citep{zhang2020bertscore} and encoder-based semantic reward modeling~\citep{pappone2025encoder}. 
SCPO draws on these ideas but differs in source and granularity: its reference is an on-policy successful \emph{sibling}, not an expert demonstration or fixed text, and it rewards \emph{new} step-level progress rather than whole-path imitation or raw similarity.

\section{Method}
\label{sec:method}

SCPO is a reward-level plugin for group-based agentic RL (Figure~\ref{fig:pipeline}): it leaves the rollout collector, policy objective, and advantage estimator untouched, and acts only on the step rewards that enter advantage estimation. 
After fixing notation (\S\ref{sec:method-prelim}), SCPO proceeds in three stages. 
Given a rollout group, it takes a successful sibling as a \emph{success reference} for the failed ones (\S\ref{sec:method-anchor}). 
It then scores every failed step against the reference steps with a frozen cross-encoder, yielding a semantic similarity matrix, and from this matrix credits a failed step only when its match is both strong and reaches a part of the reference not yet credited, so that only \emph{new} progress is rewarded (\S\ref{sec:method-matcher}; the order in which steps claim the reference is set in \S\ref{sec:method-order}). 
Finally, the recovered credit is added to that step's GiGPO return before advantage estimation (\S\ref{sec:method-integration}). 
Because credit is granted for behaviour a failed step shares with a successful sibling, SCPO narrows the gap between near-identical behaviour in successful and failed rollouts, which motivates the name \emph{semantic consistency}. 
Pseudocode and full matcher details are in Appendix~\ref{app:algorithm-bound}.

\subsection{Preliminaries and Motivation}
\label{sec:method-prelim}

An LLM agent $\pi_\theta$ solves a task $x\sim p(X)$ through multi-turn interaction: at step $t$ it observes a state $s_t$, emits a textual action $a_t\sim\pi_\theta(\cdot\mid s_t,x)$, and receives a reward $r_t$ and next state $s_{t+1}$, producing a trajectory $\tau=(s_t,a_t,r_t)_{t=1}^{T}$. We focus on the sparse-reward regime, where the environment returns a task outcome $R(\tau)=\sum_t r_t$ only at the end of an episode, labelling $\tau$ a success or a failure.

Following group-based agentic RL, for each task we sample a group of $n$ trajectories $\{\tau_i\}_{i=1}^{n}$ from identical initial states and estimate advantages from group-internal statistics, using GiGPO~\citep{feng2025gigpo} as the base estimator. GiGPO forms a two-level relative advantage. At the \emph{episode level} it normalizes trajectory returns within the group,
\begin{equation}
\label{eq:gigpo-ae}
A_E(\tau_i)=\frac{R(\tau_i)-\operatorname{mean}\big(\{R(\tau_j)\}_{j=1}^{n}\big)}{F_{\mathrm{norm}}\big(\{R(\tau_j)\}_{j=1}^{n}\big)},
\end{equation}
with $F_{\mathrm{norm}}$ the group standard deviation. At the \emph{step level} it groups time steps that revisit the same environment (\emph{anchor}) state $\tilde s$, $G_S(\tilde s)=\{(a^{(i)}_t,R^{(i)}_t):s^{(i)}_t=\tilde s\}$, assigns each step its discounted return $R^{(i)}_t=\sum_{k\ge t}\gamma^{\,k-t} r^{(i)}_k$, and normalizes within the anchor-state group,
\begin{equation}
\label{eq:gigpo-as}
A_S\big(a^{(i)}_t\big)=\frac{R^{(i)}_t-\operatorname{mean}\big(\{R^{(j)}_t\}_{G_S(\tilde s)}\big)}{F_{\mathrm{norm}}\big(\{R^{(j)}_t\}_{G_S(\tilde s)}\big)} .
\end{equation}
The two signals combine into a single group-in-group advantage,
\begin{equation}
A\big(a^{(i)}_t\big)=A_E(\tau_i)+\omega\,A_S\big(a^{(i)}_t\big),\label{eq:gigpo-adv}
\end{equation}
with step-advantage weight $\omega$, optimized with the standard clipped policy-gradient objective and KL regularization~\citep{schulman2017ppo}. SCPO acts one level earlier, adding credit to the step-level discounted returns $R^{(i)}_t$ of failed trajectories before this estimator is applied (\S\ref{sec:method-integration}).

\begin{figure}[t]
\centering
\includegraphics[width=\columnwidth]{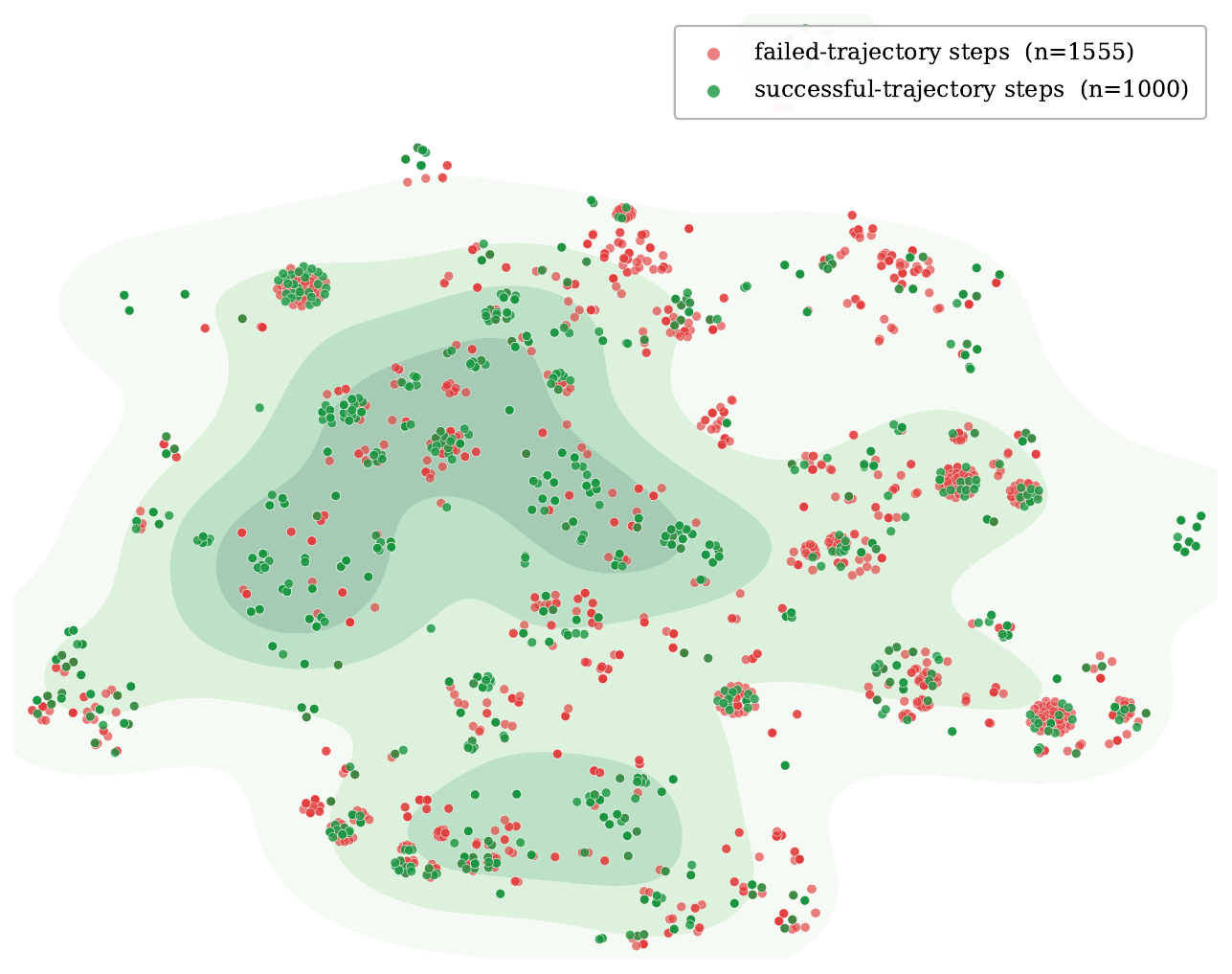}
\caption{\textbf{Failed trajectories contain successful-like steps.} Many failed-trajectory steps (red) fall inside the regions occupied by successful-trajectory steps (green), yet trajectory-level credit assigns them the same negative advantage. Embedding and construction details are in Appendix~\ref{app:semantic-neighbourhood}.}
\label{fig:motiv-tsne}
\end{figure}

\paragraph{Motivation.}
Both levels tie a step's credit to its trajectory's outcome: $A_E(\tau_i)$ is constant across a trajectory, and the step return $R^{(i)}_t$ inherits the sparse terminal reward, so every step of a failed sibling receives the same negative advantage---regardless of whether it was locally correct. Yet failed trajectories routinely contain steps near-identical to successful behaviour. Figure~\ref{fig:motiv-tsne} embeds filtered steps from early-training rollout groups and projects them to 2D: many failed-trajectory steps (red) fall inside the regions occupied by successful-trajectory steps (green shading), i.e.\ failed rollouts often make locally-successful progress that outcome-based credit nonetheless penalises. 
SCPO targets exactly these steps, recovering partial step-level credit for the successful-like behaviour inside failed trajectories.

\subsection{Success reference}
\label{sec:method-anchor}

For each group we partition its $n$ trajectories by terminal outcome into successful siblings $\mathcal{S}=\{i:R(\tau_i)>\tau_{\mathrm{succ}}\}$ and failed siblings $\mathcal{W}$ ($\tau_{\mathrm{succ}}=0$ for binary rewards), and SCPO acts only when $\mathcal{S}\neq\varnothing$. We take as the \emph{success reference} the longest successful trajectory,
\[
\tau^\star=\operatorname*{arg\,max}_{i\in\mathcal{S}}\,|\tau_i|,
\]
which supplies the largest set of reusable reference behaviours (ablated in \S\ref{sec:exp-ablate}); $\tau^\star$ is an on-policy sibling for the same task, not external supervision. Each step is encoded as a text string $x_t$ formed from its state observation $s_t$ and action $a_t$ (Figure~\ref{fig:pipeline}), so that context-dependent actions remain distinguishable; the exact format is given in Appendix~\ref{app:step-filtering}. After removing invalid and no-op steps (Appendix~\ref{app:step-filtering}), we obtain a filtered reference $\bar\tau^\star=(x^\star_1,\dots,x^\star_m)$ of length $m$ and, for each failed sibling $w$, a filtered step sequence $\bar\tau_w=(x^w_1,\dots,x^w_\ell)$ of length $\ell$, where $x^\star_u$ and $x^w_v$ are the per-step texts defined above.

\subsection{Semantic similarity and progress credit}
\label{sec:method-matcher}

\paragraph{Similarity matrix.}
SCPO scores every filtered failed step $v$ against every filtered reference step $u$ with a frozen cross-encoder $f$, giving a similarity matrix
\[
M\in[0,1]^{m\times\ell},\qquad M_{u,v}=f(x^\star_u,\,x^w_v).
\]

\paragraph{Monotonic semantic credit.}
Raw similarity is not rewarded directly: over long horizons, repeated observations, navigation templates, and retries produce many high-similarity matches that carry no new task progress. SCPO instead rewards only \emph{new} progress along the reference, governed by two rules. A failed step earns credit at a reference position $j$ only if (i)~their similarity exceeds a threshold $\theta$, and (ii)~that position lies beyond every position already credited, so credit moves strictly forward through the reference and each position is used at most once. A step meeting both rules receives soft credit
\[
r^{\mathrm{SCPO}}_{w,v}=\frac{\max\!\left(0,\,M_{j,v}-\lambda\right)}{1-\lambda},
\]
with soft base $\lambda$; any other step---sub-threshold, or re-matching an already-credited position---receives $r^{\mathrm{SCPO}}_{w,v}=0$. Thus $\theta$ sets which matches are eligible and $\lambda$ scales the credit among them. Because each reference position is credited at most once, repeated or templated behaviour cannot accumulate reward, regardless of the failed trajectory's length or how often a behaviour recurs (\S\ref{app:bound}). We give the matching procedure and pseudocode in Appendix~\ref{app:algorithm-bound}.

\subsection{Allocation order}
\label{sec:method-order}

Because credit moves strictly forward through the reference (\S\ref{sec:method-matcher}), it is a scarce, ordered resource: once a failed step claims reference position $j$, earlier positions close to that trajectory's remaining steps, so the processing order $\pi$ decides which steps claim the limited reference positions. Temporal order $\pi=(1,\dots,\ell)$ has a structural failure mode---a failed sibling's early steps are dominated by templated navigation shared by nearly all rollouts regardless of outcome, so crediting front-to-back spends the limited positions on these low-discriminative early steps. SCPO instead \emph{reorders} this processing order so that steps from all positions compete on equal footing to extend the credited progress, rather than being credited strictly front-to-back. Such an order is available at no extra cost from the training pipeline (Appendix~\ref{app:matcher}). This permutation only changes \emph{which} step first claims a position---not the trajectory, the reward timestamp to which credit is written back, or the discounted returns---thereby ensuring that steps from all positions are matched fairly. We analyse the resulting gain over temporal order in \S\ref{sec:exp-allocation-order}.

\subsection{Integration with the base trainer}
\label{sec:method-integration}

SCPO adds the recovered credit directly to the step-level discounted return of failed siblings,
\begin{align}
\tilde R^{(i)}_t&=R^{(i)}_t+\alpha\,r^{\mathrm{SCPO}}_{i,t},\notag\\
\tilde A_S\big(a^{(i)}_t\big)&=A_S\big(a^{(i)}_t\big)\Big|_{R\to\tilde R},\label{eq:sapo-shaped-as}
\end{align}
where $R^{(i)}_t=\sum_{k\ge t}\gamma^{\,k-t}r^{(i)}_k$ is GiGPO's discounted step return and $r^{\mathrm{SCPO}}_{i,t}\ge 0$ is the monotonic semantic credit assigned to step $t$ of sibling $i$ by the matcher of \S\ref{sec:method-matcher}, with $r^{\mathrm{SCPO}}_{i,t}=0$ for successful siblings, filtered steps, and steps that do not reach a new reference position. The credit is added at the step where it is earned rather than propagated back through the discount, following GiGPO's convention for step-level reward terms such as its invalid-action penalty. The shaped return enters the estimator only through the step-level advantage: $\tilde A_S(a^{(i)}_t)$ is Eq.~\eqref{eq:gigpo-as} evaluated on $\tilde R^{(i)}_t$, giving the shaped advantage $\tilde A(a^{(i)}_t)=A_E(\tau_i)+\omega\,\tilde A_S(a^{(i)}_t)$. The episode advantage $A_E$ is computed from the unshaped task outcome $R(\tau_i)$ in Eq.~\eqref{eq:gigpo-ae} and is deliberately left intact: credit is redistributed \emph{within} a failed trajectory but never enters its episode return or its success/failure label, so SCPO cannot promote a failure to a success. SCPO therefore keeps the GiGPO objective and optimizer and only substitutes $\tilde A$ for $A$. The entire intervention is thus the advantage shift
\begin{equation}
\label{eq:sapo-delta}
\tilde A\big(a^{(i)}_t\big)-A\big(a^{(i)}_t\big)=\omega\big(\tilde A_S(a^{(i)}_t)-A_S(a^{(i)}_t)\big),
\end{equation}
nonzero only for failed-sibling steps whose discounted return is shifted by recovered credit (and, through anchor-state normalization, for their step-group peers): SCPO reweights the policy gradient toward failed steps that make new semantic progress. Its sole overhead is one frozen cross-encoder pass over already-collected steps---no learned critic, process reward model, verifier, demonstrations, or extra rollouts.

\section{Experiments}
\label{sec:experiments}

In this section, we present empirical evaluations of SCPO on challenging agentic benchmarks. Specifically, we aim to demonstrate: (1) the effectiveness of SCPO in training strong LLM agents and its competitiveness with state-of-the-art group-based baselines; (2) where the gains arise, in particular their concentration in the harder multi-step task families; (3) the contribution of SCPO's design choices, with a focus on allocation order, through ablation studies; and (4) the computational overhead introduced by the frozen cross-encoder shaper.

\begin{table*}[!t]
\centering
\scriptsize
\setlength{\tabcolsep}{2pt}
\renewcommand{\arraystretch}{1.05}
\begin{tabular*}{\textwidth}{@{\extracolsep{\fill}} l l ccccccc cc @{}}
\toprule
& & \multicolumn{7}{c}{\textbf{ALFWorld}} & \multicolumn{2}{c}{\textbf{WebShop}} \\
\cmidrule(lr){3-9}\cmidrule(lr){10-11}
\textbf{Type} & \textbf{Method} & Pick & Look & Clean & Heat & Cool & Pick2 & \textbf{All} & Score & \textbf{Succ.} \\
\midrule
\multicolumn{11}{l}{\textit{Closed-source LLMs}} \\
Prompting & GPT-4o~\citep{achiam2023gpt4} & 75.3 & 60.8 & 31.2 & 56.7 & 21.6 & 49.8 & 48.0 & 31.8 & 23.7 \\
Prompting & Gemini-2.5-Pro~\citep{geminiteam2024gemini} & 92.8 & 63.3 & 62.1 & 69.0 & 26.6 & 58.7 & 60.3 & 42.5 & 35.9 \\
\midrule
\multicolumn{11}{l}{\textit{Qwen2.5-1.5B-Instruct}} \\
Prompting & Qwen2.5 & 5.9 & 5.5 & 3.3 & 9.7 & 4.2 & 0.0 & 4.1 & 23.1 & 5.2 \\
Prompting & ReAct~\citep{yao2023react} & 17.4 & 20.5 & 15.7 & 6.2 & 7.7 & 2.0 & 12.8 & 40.1 & 11.3 \\
Prompting & Reflexion~\citep{shinn2023reflexion} & 35.3 & 22.2 & 21.7 & 13.6 & 19.4 & 3.7 & 21.8 & 55.8 & 21.9 \\
RL & PPO & 64.8$_{\pm 3.5}$ & 40.5$_{\pm 6.9}$ & 57.1$_{\pm 4.9}$ & 60.6$_{\pm 6.6}$ & 46.4$_{\pm 4.0}$ & 47.4$_{\pm 1.9}$ & 54.4$_{\pm 3.1}$ & 73.8$_{\pm 3.0}$ & 51.5$_{\pm 2.9}$ \\
RL & RLOO & 88.3$_{\pm 3.0}$ & 52.8$_{\pm 8.6}$ & 71.0$_{\pm 5.9}$ & 62.8$_{\pm 8.7}$ & 66.4$_{\pm 5.5}$ & 56.9$_{\pm 4.7}$ & 69.7$_{\pm 2.5}$ & 73.9$_{\pm 5.6}$ & 52.1$_{\pm 6.7}$ \\
RL & GRPO & 85.3$_{\pm 1.5}$ & 53.7$_{\pm 8.0}$ & 84.5$_{\pm 6.8}$ & 78.2$_{\pm 7.9}$ & 59.7$_{\pm 5.0}$ & 53.5$_{\pm 5.6}$ & 72.8$_{\pm 3.6}$ & 75.8$_{\pm 3.5}$ & 56.8$_{\pm 3.8}$ \\
RL & EMPG$^{*}$ & 85.5 & 33.5 & 78.9 & 76.2 & 74.7 & 69.1 & 73.7 & 80.4 & 60.8 \\
RL & GiGPO$^{*}$ & 94.4$_{\pm 5.9}$ & 67.5$_{\pm 4.6}$ & 94.8$_{\pm 3.8}$ & 94.4$_{\pm 7.8}$ & 79.8$_{\pm 4.7}$ & 76.4$_{\pm 5.4}$ & 86.7$_{\pm 1.7}$ & 83.1$_{\pm 1.6}$ & 65.0$_{\pm 3.2}$ \\
RL & HCAPO$^{*}$ & 88.6$_{\pm 7.0}$ & 75.0$_{\pm 0.0}$ & \textbf{97.6}$_{\pm 1.8}$ & 90.7$_{\pm 6.9}$ & 84.2$_{\pm 0.0}$ & 74.2$_{\pm 6.9}$ & 87.0$_{\pm 4.1}$ & 83.8$_{\pm 0.7}$ & 68.5$_{\pm 1.0}$ \\
RL & HGPO ($K\!=\!2$, 160 it.)$^{*}$ & -- & -- & -- & -- & -- & -- & 92.77$_{\pm 1.08}$ & 85.56$_{\pm 2.86}$ & 71.54$_{\pm 4.00}$ \\
RL & \textbf{SCPO (ours)} & \textbf{96.0}$_{\pm 4.4}$ & \textbf{88.6}$_{\pm 12.7}$ & 93.4$_{\pm 8.2}$ & \textbf{96.5}$_{\pm 3.1}$ & \textbf{92.8}$_{\pm 12.5}$ & \textbf{90.9}$_{\pm 3.4}$ & \textbf{93.7}$_{\pm 4.1}$ & \textbf{89.3}$_{\pm 1.8}$ & \textbf{74.8}$_{\pm 2.0}$ \\
\midrule
\multicolumn{11}{l}{\textit{Qwen2.5-7B-Instruct}} \\
Prompting & Qwen2.5 & 33.4 & 21.6 & 19.3 & 6.9 & 2.8 & 3.2 & 14.8 & 26.4 & 7.8 \\
Prompting & ReAct~\citep{yao2023react} & 48.5 & 35.4 & 34.3 & 13.2 & 18.2 & 17.6 & 31.2 & 46.2 & 19.5 \\
Prompting & Reflexion~\citep{shinn2023reflexion} & 62.0 & 41.6 & 44.9 & 30.9 & 36.3 & 23.8 & 42.7 & 58.1 & 28.8 \\
RL & PPO & 92.3$_{\pm 4.0}$ & 64.0$_{\pm 8.4}$ & 92.5$_{\pm 2.4}$ & 89.5$_{\pm 7.0}$ & 80.3$_{\pm 2.0}$ & 68.8$_{\pm 8.3}$ & 80.4$_{\pm 2.7}$ & 81.4$_{\pm 3.1}$ & 68.7$_{\pm 5.1}$ \\
RL & RLOO & 87.6$_{\pm 4.3}$ & 78.2$_{\pm 8.3}$ & 87.3$_{\pm 5.8}$ & 81.3$_{\pm 7.6}$ & 71.9$_{\pm 5.2}$ & 48.9$_{\pm 8.4}$ & 75.5$_{\pm 4.6}$ & 80.3$_{\pm 3.2}$ & 65.7$_{\pm 4.0}$ \\
RL & GRPO & 90.8$_{\pm 5.1}$ & 66.1$_{\pm 6.7}$ & 89.3$_{\pm 5.4}$ & 74.7$_{\pm 6.9}$ & 72.5$_{\pm 5.4}$ & 64.7$_{\pm 7.3}$ & 77.6$_{\pm 5.2}$ & 79.3$_{\pm 2.8}$ & 66.1$_{\pm 3.7}$ \\
RL & EMPG$^{*}$ & 92.9 & 75.2 & 74.8 & 86.3 & 73.7 & 65.3 & 78.5 & 81.0 & 69.3 \\
RL & GiGPO$^{*}$ & 97.7$_{\pm 1.6}$ & 82.7$_{\pm 7.9}$ & \textbf{98.8}$_{\pm 1.6}$ & 83.7$_{\pm 7.2}$ & 89.3$_{\pm 8.2}$ & 79.2$_{\pm 6.6}$ & 90.8$_{\pm 1.3}$ & 84.4$_{\pm 2.9}$ & 72.8$_{\pm 3.2}$ \\
RL & HCAPO$^{*}$ & 99.1$_{\pm 1.3}$ & 90.3$_{\pm 2.0}$ & 97.3$_{\pm 1.9}$ & 81.8$_{\pm 8.8}$ & \textbf{90.8}$_{\pm 6.6}$ & 81.9$_{\pm 10.0}$ & 91.4$_{\pm 2.3}$ & 85.1$_{\pm 1.3}$ & 73.8$_{\pm 2.8}$ \\
RL & HGPO ($K\!=\!2$, 160 it.)$^{*}$ & -- & -- & -- & -- & -- & -- & \textbf{95.44}$_{\pm 0.62}$ & \textbf{88.96}$_{\pm 1.04}$ & \textbf{78.51}$_{\pm 1.40}$ \\
RL & \textbf{SCPO (ours)} & \textbf{100.0}$_{\pm 0.0}$ & \textbf{100.0}$_{\pm 0.0}$ & 97.2$_{\pm 2.4}$ & \textbf{92.9}$_{\pm 7.2}$ & 89.3$_{\pm 9.4}$ & \textbf{94.3}$_{\pm 2.6}$ & 95.3$_{\pm 1.6}$ & 88.0$_{\pm 3.8}$ & 77.5$_{\pm 3.0}$ \\
\bottomrule
\end{tabular*}
\caption{Main results on ALFWorld (\textsc{valid\_seen}, per-family and overall success rate \%) and WebShop (average score and success rate \%). SCPO wraps GiGPO and changes only the step-reward signal before advantage estimation. SCPO numbers are mean$\pm$std over three seeds; best per column in bold. $^{*}$ Results for these baselines are quoted from their original papers~\citep{empg2025,feng2025gigpo,tan2026hcapo,he2026hgpo}; HGPO does not report a per-family breakdown.}
\label{tab:main}
\end{table*}

\subsection{Setup}
\label{sec:exp-setup}

\paragraph{Benchmarks and metrics.} We evaluate on ALFWorld~\citep{shridhar2021alfworld} and WebShop~\citep{webshop}. ALFWorld contains embodied household tasks with $T_{\max}=50$; we report success rate on the \texttt{valid\_seen} split. WebShop requires web-page navigation toward an attribute-matched purchase with $T_{\max}=15$; we report average task score and binary task-success rate.

\paragraph{Baselines and implementation.} We compare with prompting-only agents, standard RL fine-tuning methods, and group-based agentic RL methods. The most important baseline is GiGPO, which SCPO directly wraps. We also include HGPO (at its $K\!=\!2$ setting) as a strong published reference point and HCAPO as a hindsight-style critic refinement method. Reported results for GiGPO, HCAPO, and HGPO are quoted directly from their respective papers~\citep{feng2025gigpo,tan2026hcapo,he2026hgpo}. For Qwen2.5-1.5B-Instruct and Qwen2.5-7B-Instruct, we follow the public \texttt{verl-agent} GiGPO recipe~\citep{feng2025gigpo} using commit \texttt{796ed31}. To make the comparison controlled, we keep our configuration \emph{strictly identical} to GiGPO and change only SCPO's step-reward shaping; the shared hyperparameters are listed in Appendix~\ref{app:hparams-train}. We note that HGPO's reported numbers instead use $160$ training iterations, i.e.\ a slightly larger training budget than the $150$ used by GiGPO, HCAPO, and SCPO. Cross-encoder scoring uses frozen BGE-Reranker-v2-m3~\citep{bgereranker}. Unless otherwise stated, we report mean$\pm$std over three random seeds. Full details are in Appendix~\ref{app:experiments}.

\subsection{Main results}
\label{sec:exp-main}

Table~\ref{tab:main} reports ALFWorld and WebShop performance. The central comparison is GiGPO versus SCPO: since SCPO changes only the step-reward signal, the GiGPO--SCPO gap isolates the effect of its reward shaping.

\paragraph{SCPO achieves state-of-the-art results at 1.5B.} At 1.5B, SCPO reaches $93.7\pm4.1$ ALFWorld success and $74.8\pm2.0$ WebShop task-success. Relative to GiGPO, the recipe SCPO wraps, this is $+7.0$ on ALFWorld and $+9.8$ on WebShop task-success, and SCPO also edges past the strongest published baseline HGPO ($92.77$ / $71.54$). Both margins are large relative to seed variance, indicating that semantic progress credit adds useful supervision on top of the group-based estimator. The analysis of SCPO's mechanism is presented separately in our ablations (\S\ref{sec:exp-ablate}).

\begin{table*}[!t]
\centering
\footnotesize
\setlength{\tabcolsep}{3pt}
\renewcommand{\arraystretch}{0.95}
\begin{tabular}{l cccccc c}
\toprule
\textbf{Variant} & Pick & Look & Clean & Heat & Cool & Pick2 & \textbf{All} \\
\midrule
\textbf{SCPO} \emph{(default)} & 96.0$_{\pm 4.4}$ & \textbf{88.6}$_{\pm 12.7}$ & 93.4$_{\pm 8.2}$ & 96.5$_{\pm 3.1}$ & 92.8$_{\pm 12.5}$ & \textbf{90.9}$_{\pm 3.4}$ & \textbf{93.7}$_{\pm 4.1}$ \\
\quad chronological order & 97.0$_{\pm 3.0}$ & 82.8$_{\pm 0.9}$ & 89.0$_{\pm 3.9}$ & 95.8$_{\pm 7.2}$ & 86.0$_{\pm 14.0}$ & 85.2$_{\pm 7.8}$ & 90.4$_{\pm 2.0}$ \\
\quad $-$ monotonicity & \textbf{100.0}$_{\pm 0.0}$ & 66.5$_{\pm 10.4}$ & \textbf{95.7}$_{\pm 2.1}$ & \textbf{100.0}$_{\pm 0.0}$ & 85.7$_{\pm 9.3}$ & 77.8$_{\pm 12.3}$ & 90.1$_{\pm 1.6}$ \\
\quad shortest reference & \textbf{100.0}$_{\pm 0.0}$ & 81.5$_{\pm 8.5}$ & 93.3$_{\pm 6.7}$ & 93.8$_{\pm 6.3}$ & \textbf{94.7}$_{\pm 2.3}$ & 80.5$_{\pm 10.2}$ & 91.7$_{\pm 2.7}$ \\
\bottomrule
\end{tabular}
\caption{Ablation of SCPO's three design choices on ALFWorld 1.5B. Each row changes one choice from the default SCPO---reordered processing, longest successful reference, and monotonicity---holding all else fixed: \emph{chronological order} processes failed steps in their original temporal order; \emph{$-$ monotonicity} removes the monotonic credit rule so already-credited reference positions can be rewarded again (repeated credit); \emph{shortest reference} uses the shortest rather than the longest successful sibling. All values are mean$\pm$std over three random seeds.}
\label{tab:abl-kmp}
\end{table*}

\paragraph{Gains concentrate on the hardest task families.} The per-family columns of Table~\ref{tab:main} show where this improvement comes from. On the families where GiGPO is already near-saturated (\textsc{Pick}, \textsc{Clean}, \textsc{Heat}), SCPO is comparable, staying within a few points. The largest gains instead concentrate in the harder multi-step families: at 1.5B, SCPO improves \textsc{Look} from $67.5$ to $88.6$ ($+21.1$), \textsc{Cool} from $79.8$ to $92.8$ ($+13.0$), and \textsc{Pick2} from $76.4$ to $90.9$ ($+14.5$) over GiGPO. These are exactly the families where a failed rollout is most likely to make partial progress before failing---navigating to the wrong receptacle, cleaning the wrong object, or completing only one of two required placements---so that the trajectory's failure signal overrides the credit for genuinely useful intermediate behaviour. SCPO's semantic progress credit has the most to recover precisely here, whereas single-step families leave little partial progress to credit.

\paragraph{SCPO remains among the strongest methods at 7B.} At 7B, SCPO remains the strongest or near-strongest method, reaching $95.3$ on ALFWorld (vs.\ GiGPO $90.8$, on par with HGPO $95.44$) and $77.5$ WebShop task-success (vs.\ GiGPO $72.8$). The same per-family pattern holds, with SCPO reaching $100$ on \textsc{Look} and $94.3$ on \textsc{Pick2}. The absolute headroom shrinks as the base policy strengthens, consistent with SCPO's intended role: it is most useful when failed rollouts still contain recoverable successful-like local behaviour, which is especially common for weaker or mid-training policies.

\begin{figure*}[t]
\centering
\includegraphics[width=\textwidth]{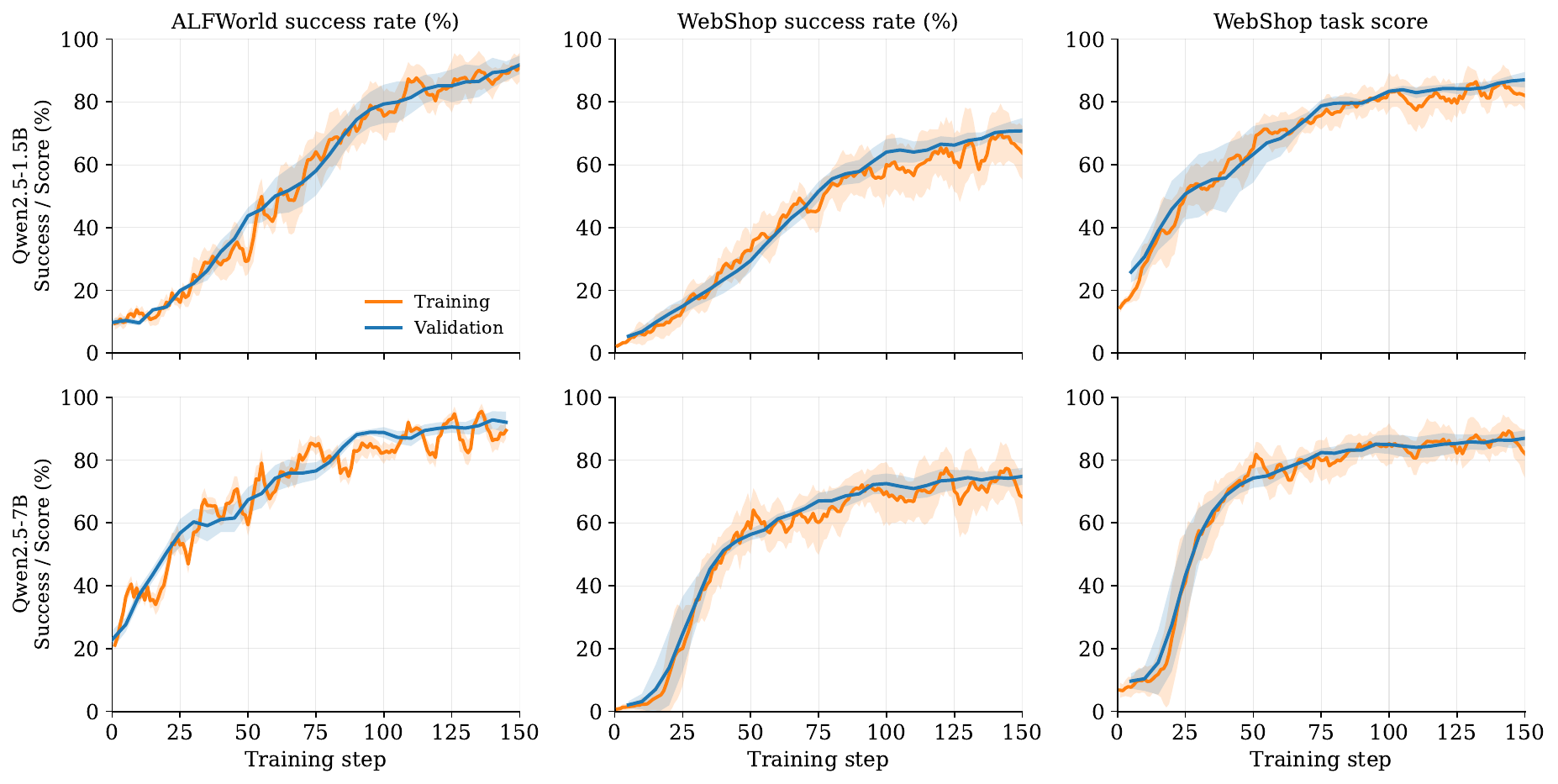}
\caption{\textbf{Learning dynamics over 150 training steps} for Qwen2.5-1.5B (top) and Qwen2.5-7B (bottom). Each panel shows SCPO's training and validation curves (three-seed mean, shaded $\pm$std) for ALFWorld success rate, WebShop success rate, and WebShop task score. Validation rises steadily and plateaus within the budget, with training closely tracking it across both benchmarks and model sizes.}
\label{fig:learning-dynamics}
\end{figure*}

Figure~\ref{fig:learning-dynamics} shows the full learning dynamics over training: across both benchmarks and model sizes, validation success and task score rise steadily and plateau within the $150$-step budget, with training closely tracking validation.

\subsection{Ablations}
\label{sec:exp-ablate}

We next isolate which design choices are responsible for stable semantic progress credit. All ablations are run on ALFWorld with Qwen2.5-1.5B-Instruct, changing one design dimension at a time. All variants, including the ablations, are averaged over three seeds.

\subsubsection{Allocation order matters}
\label{sec:exp-allocation-order}

SCPO does not impose a chronological prior on credit allocation (\S\ref{sec:method-order}). We test whether this matters by replacing the reordered allocation with strict chronological matching---processing failed steps in their original temporal order---while holding every other component fixed.

The reordered allocation outperforms chronological matching by $+3.3$ points (Table~\ref{tab:abl-kmp}, \emph{chronological order} row). By letting every failed step compete for the reference on equal footing rather than crediting front-to-back, the reordered order matches steps to the reference more fairly, raising both the number of credited steps per failed trajectory and the mean auxiliary reward over training (Figure~\ref{fig:alloc-credit-trend}). As discussed in \S\ref{sec:method-order}, temporal order instead spends the limited reference positions on early, low-discriminative templated steps; letting steps from all positions compete surfaces the genuinely useful behaviour that tends to appear later or interleaved with detours, which is where SCPO's gains concentrate. We read this as evidence that strict temporal priority is a weaker allocation rule for failed agentic trajectories than order-agnostic competition.

\subsubsection{Monotonic crediting prevents reward inflation}
\label{sec:exp-budget}

We next test whether SCPO's monotonic credit rule is necessary. The monotonic rule emits credit only when a match reaches a reference position beyond those already credited. Without this rule, repeated template matches can be converted into reward multiple times, even when the failed trajectory is revisiting similar observations rather than making new task progress.

Table~\ref{tab:abl-kmp} shows that removing the monotonicity constraint lowers overall success to $90.1$ ($-3.6$ relative to SCPO). This supports the distinction between semantic progress and semantic repetition: a failed trajectory may repeatedly match the same successful-like template, but repeatedly rewarding such matches inflates auxiliary reward without reflecting new task progress.

\subsubsection{Longer references preserve exploration}
\label{sec:exp-reference}

SCPO selects the longest successful sibling as the reference (\S\ref{sec:method-anchor}). Intuitively, the \emph{shortest} successful sibling---the most efficient solution---might seem the cleaner, more refined reference. Yet Table~\ref{tab:abl-kmp} shows it lowers overall success to $91.7$ ($-2.0$ relative to SCPO), with the drop concentrated in the harder multi-step families (\textsc{Look}, \textsc{Pick2}). A short reference consists almost entirely of the core steps that nearly every rollout shares, so crediting against it concentrates reward on this single efficient path; this may suppress policy entropy and discourage exploration, hurting precisely the multi-step tasks that require longer or more varied solutions. The longest reference instead exposes more intermediate and later milestones, giving failed steps a wider and more task-specific set of positions to advance against, which better preserves exploration.

\subsection{Mechanism and cost}
\label{sec:exp-cost}

Diagnostics in Appendix~\ref{app:mechanism-evidence} support the intended mechanism. The neighbourhood analysis (Appendix~\ref{app:semantic-neighbourhood}) shows that the failed steps SCPO credits coincide tightly with successful-sibling behaviour---most are near-exact matches---so SCPO recovers a substantial amount of useful signal hidden in failed rollouts rather than applying a uniform bonus. The credit-spectrum heatmaps (Appendix~\ref{app:case-study}) show this failed-versus-reference overlap growing denser from early to late checkpoints, so as the policy improves SCPO finds progressively more reusable behaviour to credit. Finally, the single-knob sensitivity analysis (Appendix~\ref{app:hparam-sweep}) shows that performance stays stable across a range of $\theta$ and $\lambda$, indicating that SCPO is fairly robust to its hyperparameter settings rather than sensitive to a particular choice.

%The mechanism diagnostics should be interpreted as evidence of increasing semantic overlap, not causal verification of every credited step. SCPO does not prove that a credited step caused eventual success; rather, it uses successful siblings as on-policy semantic references and converts new successful-like progress into bounded auxiliary reward. This is precisely why monotonicity is important: it limits the impact of imperfect semantic matches while still extracting useful signal from failed rollouts.

SCPO adds $29.3$s per training step in the ALFWorld 1.5B setting ($10.4\%$ overhead) and increases peak GPU memory from $33.5$GB to $35.2$GB. Nearly all overhead comes from frozen cross-encoder inference; the monotonic credit matcher itself is negligible. A full timing breakdown is provided in Appendix~\ref{app:cost-detail}.

\section{Conclusion}
We identified \emph{semantic credit inconsistency} as a credit-assignment
failure mode in group-based agentic RL: because a step's credit is tied to
its trajectory's eventual outcome, a locally correct step in a rollout that
fails later is penalised, and two near-identical steps can receive opposite
training signals. To address it, we proposed \textbf{SCPO}, a value-free
reward-level plugin that recovers step credit by matching each failed step
against a successful in-group sibling and rewarding only the \emph{new}
progress it makes---crediting each reference position at most once, with no
critic, reward model, verifier, demonstrations, or extra rollouts. On
ALFWorld and WebShop, SCPO improves the GiGPO backbone it wraps, reaching
state-of-the-art results at 1.5B and remaining competitive with the
strongest published baselines at 7B, with the largest gains on the harder
multi-step families where failed rollouts carry the most recoverable
progress. Because SCPO intervenes only at the step-reward level, it is
agnostic to the underlying group-based algorithm and could be paired with
other estimators. A natural next step is to recover credit even when a group
contains no successful sibling---for example, by drawing references from a
buffer of past successes---so that SCPO also helps in the all-failure regime
of early or extremely sparse training.

\section*{Limitations}
\label{sec:limitations}

% \paragraph{Evaluation scope and OOD generalisation.}
% Our complete three-seed results cover ALFWorld and WebShop with
% Qwen2.5-1.5B-Instruct and Qwen2.5-7B-Instruct. These benchmarks
% represent embodied and web-navigation agents, but do not cover
% open-ended research agents, code-generation agents, large tool-use
% spaces, or robotic control. We also report only in-distribution
% success and leave a systematic out-of-distribution evaluation (e.g.,
% ALFWorld \texttt{valid\_unseen}) to future work.

% \paragraph{Single-seed mechanism studies.}
% The main SCPO results are averaged over three seeds, but the ablation
% rows in Tab.~\ref{tab:abl-kmp} and the sensitivity sweep in
% Figure~\ref{fig:hparam-sweep} are single-seed due to compute
% constraints. They should be interpreted as mechanism checks for
% reference-slot budgeting and reference selection, not as precise estimates
% of average effects.

\paragraph{Dependence on successful siblings.}
SCPO can recover credit only in groups that contain at least one
successful sibling. When all rollouts fail, the group is left
unchanged. This makes SCPO complementary to trajectory-level
group-based RL rather than a replacement for it; in extremely sparse
or early-training regimes, its benefit may appear only after the base
policy occasionally solves the task.

\paragraph{Approximate semantic matching, not exact verification.}
SCPO credits a failed step by semantic similarity against a successful
sibling rather than by exact state--action equivalence. The frozen
cross-encoder returns a soft alignment score, so the matcher operates
on fuzzy semantic overlap and can both over- and under-credit: two
steps with similar surface form may be scored as matching even when
their effects differ, while genuinely equivalent steps phrased
differently may be missed. This is well suited to embodied and
web-navigation tasks such as ALFWorld and WebShop, where progress is
reflected in coarse, observable state changes (locations, objects,
page transitions) that a general-purpose reranker captures reliably.
It is a poorer fit for domains that require exact, symbolic matching:
in preliminary experiments applying the same recipe to reasoning RL on
mathematics and code datasets, we observed no clear improvement, which
we attribute to the off-the-shelf cross-encoder---never trained with a
contrastive objective on such content---producing only a coarse
similarity signal that cannot reliably separate a correct intermediate
derivation step from a subtly incorrect one. Adapting the scorer to
the target domain (e.g., a contrastively trained or verifier-based
matcher) is a natural extension that we leave to future work.

\paragraph{Local semantic credit is not causal verification.}
SCPO avoids forcing failed trajectories to imitate the exact temporal
path of a successful rollout, but it also does not verify that matched
local behaviours form a globally coherent solution. A failed step may
semantically resemble a success reference slot while being irrelevant or
harmful in context. Crediting each reference position at most once limits the
magnitude of such errors, but does not guarantee policy
improvement; the realised update still depends on advantage
normalisation, PPO clipping, KL regularisation, and optimisation
dynamics.

% \clearpage
\bibliography{main}

\clearpage
\newpage
\appendix

\section{Experimental Details}
\label{app:experiments}

This appendix provides the experimental details supporting \S\ref{sec:experiments}. We include benchmark protocols, training hyperparameters, SCPO shaper settings, evaluation details, and compute configuration.

\subsection{Benchmarks}
\label{app:benchmarks}

\paragraph{ALFWorld.} ALFWorld~\citep{shridhar2021alfworld} is an embodied-instruction benchmark built on text-based household tasks aligned with the visual ALFRED environment~\citep{shridhar2020alfred}. Each episode starts from a natural-language goal and a textual observation of the agent's current location. 
%The agent issues actions such as \texttt{go to fridge 1}, \texttt{open fridge 1}, \texttt{take apple 1 from fridge 1}, and \texttt{put apple 1 in sidetable 1}; the environment returns a textual observation after each action. We follow the standard six task families: \textsc{pick\_and\_place}, \textsc{pick\_clean\_then\_place}, \textsc{pick\_heat\_then\_place}, \textsc{pick\_cool\_then\_place}, \textsc{pick\_two\_obj\_and\_place}, and \textsc{look\_at\_obj\_in\_light}. 
We evaluate on \textsc{valid\_seen}. The maximum episode length is $T_{\max}=50$, matching the GiGPO ALFWorld configuration~\citep{feng2025gigpo}.

\paragraph{WebShop.} WebShop~\citep{webshop} is a simulated e-commerce environment where the agent must purchase a product matching a natural-language description. 
Actions include search, click, and purchase commands, and observations correspond to rendered product pages. 
Rewards are delivered after the purchase action. 
We use the standard protocol with $T_{\max}=15$ and report both the average task score and the binary task-success rate, matching the GiGPO WebShop configuration~\citep{feng2025gigpo}.

\subsection{Hardware and software}
\label{app:hardware}

All experiments are run on a single node with $4\times$NVIDIA A800 80GB GPUs, 32 CPU cores, and approximately 2~TiB of system memory. 
We use the public \texttt{verl-agent} training stack~\citep{feng2025gigpo} at commit \texttt{796ed31}. 
Inference is served by vLLM with FlashAttention~\citep{flashattention} enabled. Training uses FSDP~\citep{fsdp2zhao} for actor and reference-policy sharding.
Cross-encoder scoring uses BGE-Reranker-v2-m3~\citep{bgereranker}, executed in a separate subprocess sharing GPU~0.

Every run trains for $150$ iterations on the $4\times$A800 GPUs above. A single SCPO run takes approximately $13$\,h (ALFWorld) and $7$\,h (WebShop) with Qwen2.5-1.5B-Instruct, and approximately $36$\,h (ALFWorld) and $27$\,h (WebShop) with Qwen2.5-7B-Instruct; wall-clock times vary with cluster load. Across three seeds per setting, the SCPO runs reported here total on the order of $1{,}000$ GPU-hours. As detailed in \S\ref{sec:exp-cost}, SCPO's frozen cross-encoder adds only ${\approx}10\%$ to GiGPO's per-step time, so this cost is dominated by the underlying multi-turn rollouts rather than by SCPO's shaping.

\subsection{Training hyperparameters}
\label{app:hparams-train}

Table~\ref{tab:hparams} lists the training hyperparameters used in our SCPO runs. To keep the comparison controlled, our configuration is \emph{strictly identical} to GiGPO, and the only change relative to GiGPO is SCPO's step-reward shaping. WebShop uses the same configuration except for the environment horizon and benchmark-specific reward scale.

\begin{table}[t]
\small
\centering
\begin{tabular}{ll}
\toprule
\textbf{Hyperparameter} & \textbf{Value} \\
\midrule
Base model                       & Qwen2.5-1.5B/7B-Instruct \\
Group size $n$                   & 8 \\
Train batch size                 & 16 tasks / step \\
Validation batch size            & 128 episodes \\
PPO mini-batch size              & 256 \\
PPO micro-batch / GPU            & 32 (1.5B) / 8 (7B) \\
Learning rate                    & $1\times10^{-6}$ \\
Optimizer                        & AdamW \\
KL coefficient                   & 0.01 \\
Discount $\gamma$                & 0.95 \\
Invalid-action penalty           & 0.1 \\
Max prompt length                & 2048 \\
Max response length              & 512 \\
History context $K$              & 2 \\
Training iterations              & 150 \\
Save / validation frequency      & every 5 steps \\
Rollout temperature              & 1.0 \\
Validation temperature           & 0.4 \\
Random seeds                     & $\{0,1,2\}$ \\
\bottomrule
\end{tabular}
\caption{Training hyperparameters.}
\label{tab:hparams}
\end{table}

\subsection{SCPO shaper configuration}
\label{app:hparams-skmp}

Table~\ref{tab:hparams-skmp} lists the SCPO-specific settings. The default shaper uses monotonic semantic credit with reordered processing. The chronological ablation changes only the processing order by setting \texttt{balance\_batch=False}.

\begin{table}[t]
\centering
\footnotesize
\setlength{\tabcolsep}{2pt}
\renewcommand{\arraystretch}{1.05}
\newcolumntype{Y}{>{\raggedright\arraybackslash}X}
\begin{tabularx}{\columnwidth}{@{}YcY@{}}
\toprule
\textbf{Parameter} & \textbf{Symbol} & \textbf{Value} \\
\midrule
Match threshold & $\theta$ & 0.6 \\
Soft base & $\lambda$ & 0.4 \\
Aux. weight, ALFWorld & $\alpha$ & 0.5 \\
Aux. weight, WebShop & $\alpha$ & 1.0 \\
Success threshold, ALFWorld & $\tau_{\mathrm{succ}}$ & 0.0 \\
Success threshold, WebShop & $\tau_{\mathrm{succ}}$ & 9.0 \\
Reference selection & $\Pi$ & longest success \\
Apply scope & -- & failed siblings \\
Credit semantics & -- & monotonic semantic credit \\
Default order & $\pi$ & reordered \\
Chronological ablation & $\pi$ & temporal \\
Repeated-credit ablation & -- & monotonicity disabled \\
Invalid-action filter & -- & enabled \\
Observation blacklist & -- & no-op / empty obs. \\
Step text & -- & action + observation \\
Cross-encoder & $f$ & BGE-Reranker-v2-m3 \\
Precision & -- & fp16 \\
CE batch size & -- & 64 pairs \\
\bottomrule
\end{tabularx}
\caption{SCPO shaper configuration.}
\label{tab:hparams-skmp}
\end{table}

\subsubsection{\texorpdfstring{Choice of $\alpha$ across benchmarks}{Choice of alpha across benchmarks}}
\label{app:alpha-choice}

We use $\alpha=0.5$ on ALFWorld and $\alpha=1.0$ on WebShop. This choice follows the monotonic-credit shaping scale. Because each reference position contributes credit at most once (\S\ref{app:bound}), the auxiliary signal a failed sibling can receive scales with the number of filtered successful-reference positions. ALFWorld trajectories are generally longer, while WebShop trajectories are shorter. A larger WebShop $\alpha$ therefore keeps the auxiliary signal on the same order as the task reward.

The success threshold $\tau_{\mathrm{succ}}$ also differs by benchmark because the two reward scales differ. 
ALFWorld gives a binary terminal reward, so any positive return ($\tau_{\mathrm{succ}}=0$) marks a success. 
WebShop gives a graded purchase score in $[0,10]$, so we set $\tau_{\mathrm{succ}}=9.0$ to select only (near-)complete purchases as the success reference, preventing low-quality partial successes from being used as the matching target.

\subsection{Step filtering and text construction}
\label{app:step-filtering}

Each valid step is represented as a short text string combining the agent action and the resulting observation:
\[
x_{i,t}^{(g)}
=
\texttt{action}_{i,t}^{(g)}
\;\Vert\;
\texttt{observation}_{i,t}^{(g)} .
\]
We filter invalid actions and degenerate no-op observations before cross-encoder scoring. In ALFWorld, this includes observations such as \texttt{Nothing happens.}. Filtering avoids assigning credit to malformed or rejected actions that repeat across failed rollouts. The filtered successful sibling defines the semantic reference used by the monotonic credit matcher.

\subsection{Evaluation protocol}
\label{app:eval}

Evaluation runs every 5 training steps on a fixed set of 128 episodes. 
Results are averaged over three random seeds unless otherwise stated.
For ALFWorld, success rate is measured on \texttt{valid\_seen}. For WebShop, we report both average task score and binary task-success rate.

Baseline numbers marked with $^{*}$ in Table~\ref{tab:main} are quoted from the original papers~\citep{empg2025,feng2025gigpo,tan2026hcapo,he2026hgpo}.

\section{Algorithm and Monotonic Semantic Credit}
\label{app:algorithm-bound}

\subsection{SCPO pseudocode}
\label{app:pseudocode}

Algorithm~\ref{alg:sapo} summarises SCPO as applied once per training batch.

\begin{algorithm}[t]
\caption{SCPO step-reward shaping}
\label{alg:sapo}
\small
\begin{algorithmic}[1]
\Require Rollout groups, step-level discounted returns $R$ (Eq.~\ref{eq:gigpo-as}), cross-encoder $f$, threshold $\theta$, soft base $\lambda$, weight $\alpha$, reference selector $\Pi$
\Ensure Shaped step returns $\tilde R$
\For{each rollout group $g$}
    \State $S^{(g)} \gets \{i:R(\tau_i^{(g)})>\tau_{\mathrm{succ}}\}$
    \If{$S^{(g)}=\varnothing$}
        \State continue
    \EndIf
    \State Select successful reference $\tau_g^* \gets \Pi(S^{(g)})$
    \State Build filtered step texts for $\tau_g^*$ and failed siblings
    \State Compute reference self-similarity $M^*$ for semantic backtracking
    \For{each failed sibling $\tau_w^{(g)}$}
        \State Compute $M_{u,v}=f(x^\star_u,\,x^w_v)$
        \State Choose processing order $\pi$
        \State $r^{\mathrm{SCPO}}_w \gets \textsc{SingleUseProgressMatch}(M,M^*;\theta,\lambda,\pi)$
        \State Add credit to step returns: $\tilde R_w^{(g)}\gets R_w^{(g)}+\alpha\,r^{\mathrm{SCPO}}_w$
    \EndFor
\EndFor
\State \Return $\tilde R$
\end{algorithmic}
\end{algorithm}

\subsection{Monotonic credit matcher}
\label{app:matcher}

Given a similarity matrix $M\in[0,1]^{m\times \ell}$ between a filtered successful reference of length $m$ and a filtered failed sibling of length $\ell$, SCPO computes the auxiliary signal through monotonic semantic credit matching: each reference position can be credited at most once, and credit advances strictly forward through the reference. The matcher is implemented in the style of KMP sequence matching~\citep{knuth1977kmp,cormen2009clrs}, but its credit semantics differ from classical occurrence matching: SCPO rewards new reference progress, not every repeated semantic match.

Let $\pi=(\pi_1,\ldots,\pi_\ell)$ denote the processing order over failed-sibling steps. Chronological SCPO corresponds to $\pi=(1,\ldots,\ell)$ and is implemented with \texttt{balance\_batch=False}. The reordered order used by Default SCPO is the one verl-agent's \texttt{balance\_batch} already produces: it reorders steps by sequence length to balance tokens across data-parallel ranks, reducing padding and improving throughput. SCPO is applied \emph{after} this step in the pipeline, so it consumes the length-balanced order directly and reuses it as the credit-allocation permutation, decoupling processing order from temporal position at no extra cost since the pipeline computes it regardless. This order changes only which failed steps reach new reference positions first; it does not change the monotonic credit rule.

Algorithm~\ref{alg:prefix-match} gives the matcher. The state $j$ is the current matched reference position, and $j_{\max}$ is the historical credited frontier. A failed step receives credit only when it pushes $j$ strictly beyond $j_{\max}$. Backtracking may revisit earlier reference positions, but revisits cannot emit new credit unless they later lead to frontier expansion.

\begin{algorithm}[t]
\caption{Monotonic credit matcher}
\label{alg:prefix-match}
\small
\begin{algorithmic}[1]
\Require Similarity matrix $M\in[0,1]^{m\times \ell}$, reference self-similarity $M^*$, threshold $\theta$, soft base $\lambda$, order $\pi$
\Ensure Step-level SCPO rewards $r^{\mathrm{SCPO}}\in[0,1]^\ell$
\State Build semantic KMP table $\mathrm{next}(\cdot)$ from $M^*$ using threshold $\theta$
\State Initialize $j\gets -1$, $j_{\max}\gets -1$, $r^{\mathrm{SCPO}}_v\gets 0$ for all $v$
\For{$k=1$ to $\ell$}
    \State $v\gets \pi_k$
    \While{$j\geq 0$ and $M_{j+1,v}<\theta$}
        \State $j\gets \mathrm{next}(j)$
    \EndWhile
    \If{$M_{j+1,v}\geq \theta$}
        \State $j\gets j+1$
        \If{$j>j_{\max}$}
            \State $r^{\mathrm{SCPO}}_v\gets \max(0,(M_{j,v}-\lambda)/(1-\lambda))$
            \State $j_{\max}\gets j$
        \EndIf
    \EndIf
    \If{$j=m-1$}
        \State \textbf{break}
    \EndIf
\EndFor
\State \Return $r^{\mathrm{SCPO}}$
\end{algorithmic}
\end{algorithm}

\subsection{Monotonic crediting}
\label{app:bound}

By construction the matcher credits each of the $m$ filtered reference positions at most once: a positive score $r^{\mathrm{SCPO}}_{w,v}>0$ is emitted only when the match advances the historical frontier $j_{\max}$, which increases monotonically. A repeated, templated, or high-similarity match to an already-credited position therefore cannot be rewarded again, independent of the failed-trajectory length or the processing order $\pi$. This monotonicity is what stops semantic similarity from becoming an unbounded reward source---the repeated-credit ablation in \S\ref{sec:exp-budget} removes it and degrades performance. It is a structural property of the matcher, not a policy-improvement guarantee.

\section{Additional Experiments and Results}
\label{app:mechanism-evidence}

This appendix provides additional diagnostics and results for SCPO.

\subsection{Case study}
\label{app:case-study}

\begin{figure*}[t]
\centering
\includegraphics[width=\textwidth]{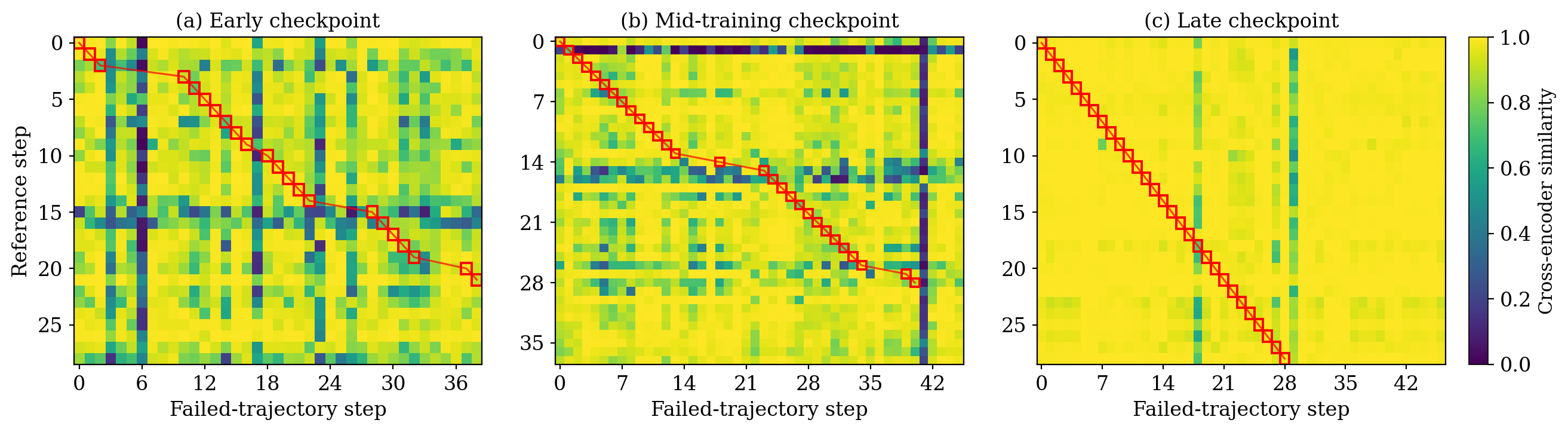}
\caption{\textbf{SCPO credit spectrum across training checkpoints} (one failed sibling vs.\ its success reference). Each cell $M_{u,v}$ is the cross-encoder similarity between reference step $u$ and failed step $v$; red boxes mark credited cells and the red line traces the forward, monotonic progress. From early to late training (left to right) the failed trajectory aligns increasingly well with the reference.}
\label{fig:spectrum}
\end{figure*}

To make the matcher concrete, Figure~\ref{fig:spectrum} visualises the cross-encoder similarity matrix between one failed sibling and its success reference at three training checkpoints. Each cell $M_{u,v}$ is the similarity between reference step $u$ and failed step $v$; red boxes mark the cells SCPO credits, and the red line traces the forward, monotonic progress through the reference. As the policy improves from early to late training (left to right), the failed trajectory's steps align increasingly well with the reference---the matrix brightens and the credited path lengthens and straightens---so SCPO finds progressively more reusable behaviour to credit. The heatmaps are displayed in trajectory order; the default reordered matcher may process steps in a different order during training, so the visualization reflects semantic-overlap density rather than the exact processing order.

\subsection{Semantic neighbourhood of the auxiliary signal}
\label{app:semantic-neighbourhood}

\begin{figure}[H]
\centering
\includegraphics[width=\columnwidth]{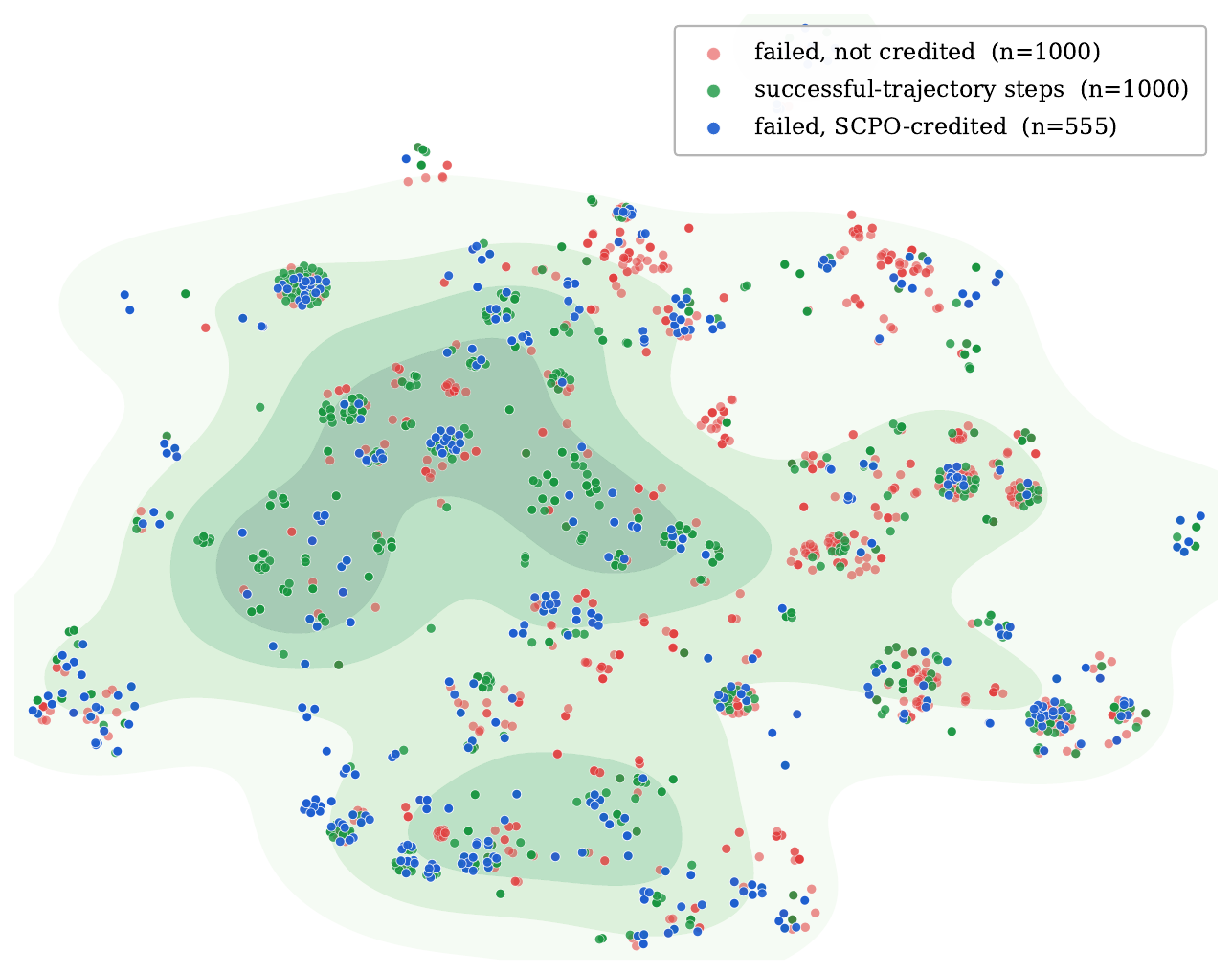}
\caption{\textbf{Semantic neighbourhood of SCPO's auxiliary signal} (ALFWorld 1.5B, five early-training batches). Each point is one filtered step embedded with all-MiniLM-L6-v2 and projected to 2D with t-SNE: successful-sibling steps (green), SCPO-credited failed steps (blue), and non-credited failed steps (red); green shading marks the density of successful-sibling steps. SCPO-credited failed steps co-locate with successful-sibling steps in tight local clusters, whereas non-credited failed steps are diffuse.}
\label{fig:semantic-tsne-scatter}
\end{figure}

To check what SCPO actually credits, we embed $2{,}555$ step strings sampled from five early-training batches with a general-purpose sentence encoder (all-MiniLM-L6-v2)~\citep{minilm} and project them to 2D with t-SNE~\citep{vandermaaten2008tsne}; each step string is the filtered \emph{(action, observation)} text defined in Appendix~\ref{app:step-filtering}. The motivation figure in the main text (Figure~\ref{fig:motiv-tsne}) shows this same 2D embedding coloured by trajectory outcome only (successful vs.\ failed steps), with green shading marking the density of successful-trajectory steps. Figure~\ref{fig:semantic-tsne-scatter} instead plots the individual steps for three groups: steps from successful in-group siblings, failed-sibling steps that received SCPO credit ($r^{\mathrm{SCPO}}>0$), and failed-sibling steps that received none. The credited-failure group coincides with the success group and is markedly tighter than the non-credited-failure group, which spreads broadly across the space. Quantitatively, the median embedding distance from a credited failed step to its nearest successful-sibling step is $0.00$ ($74\%$ are exact \emph{(action, observation)} matches) versus $0.28$ for non-credited failed steps; restricted to the non-exact subset, the medians are $0.44$ versus $0.59$ ($1.34\times$ larger). SCPO therefore credits failed steps that semantically resemble in-group successful behaviour rather than applying a uniform bonus.

Two caveats apply. The embedding uses a general-purpose encoder rather than SCPO's BGE cross-encoder (which scores pairs and does not produce standalone embeddings), so it is a faithful proxy for semantic proximity rather than SCPO's exact scoring; and the sample is drawn from early training, where most rollouts fail. The analysis should be read as evidence of semantic overlap, not as causal verification that every credited step is necessary for success.

\subsection{Hyperparameter sensitivity}
\label{app:hparam-sweep}

\begin{figure}[H]
\centering
\includegraphics[width=\columnwidth]{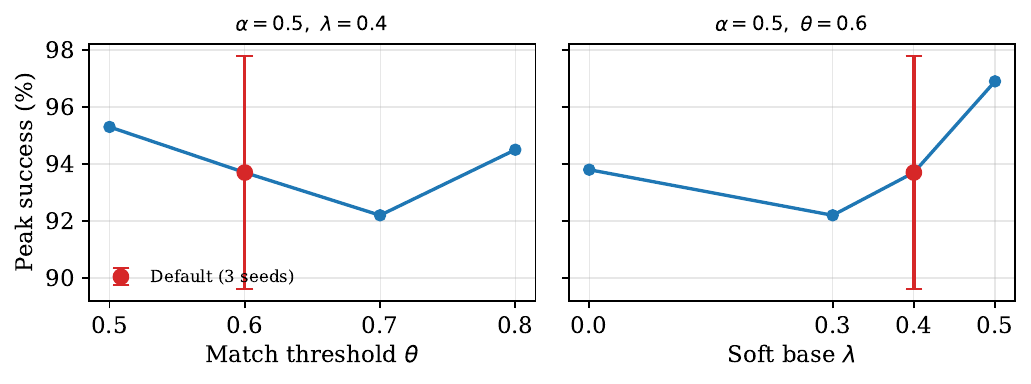}
\caption{Single-knob SCPO sensitivity on ALFWorld 1.5B. \emph{Left}: match threshold $\theta$ (at $\alpha=0.5,\lambda=0.4$). \emph{Right}: soft base $\lambda$ (at $\alpha=0.5,\theta=0.6$). All points report the best validation checkpoint reached within $150$ training steps. The red point with error bar is the default configuration (three seeds, $93.7\pm4.1$); blue points are single-seed runs.}
\label{fig:hparam-sweep}
\end{figure}

Figure~\ref{fig:hparam-sweep} reports a one-at-a-time sensitivity analysis around the default $(\alpha,\theta,\lambda)=(0.5,0.6,0.4)$ on ALFWorld with Qwen2.5-1.5B-Instruct. The default point (red) is the three-seed result $93.7\pm4.1$; all other points are single-seed runs and should be interpreted as suggestive rather than definitive.

The sweep suggests that SCPO is not overly sensitive to small changes in $\theta$ or $\lambda$, but the non-default rows are not multi-seed estimates. We therefore use this result only as supporting evidence, not as a robustness claim.

\subsection{Step-reward analysis}
\label{app:step-reward-trend}

\begin{figure}[H]
\centering
\includegraphics[width=\columnwidth]{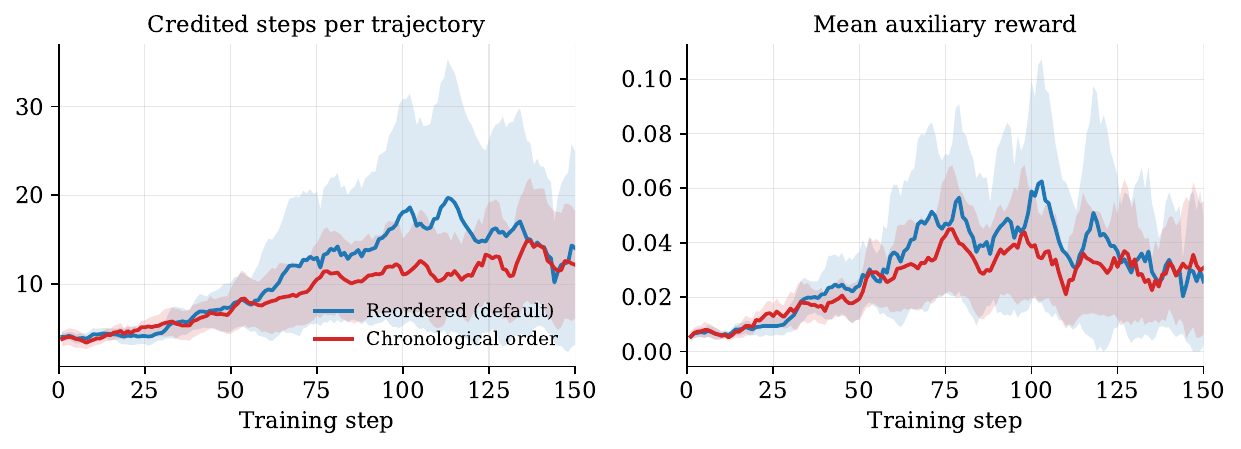}
\caption{\textbf{SCPO step-reward over training} on ALFWorld 1.5B (three-seed mean, $\pm1$ std bands, $5$-step moving average). \emph{Left}: credited steps per failed trajectory. \emph{Right}: mean auxiliary reward per step. The signal grows as the policy improves but stays bounded, and the default reordered allocation credits more than chronological order.}
\label{fig:alloc-credit-trend}
\end{figure}

Figure~\ref{fig:alloc-credit-trend} tracks the SCPO step-reward over training on ALFWorld 1.5B: the number of credited steps per failed trajectory (left) and the mean auxiliary reward per step (right), as three-seed means with $\pm1$ standard-deviation bands. Early in training the injected signal is small---few groups yet contain a successful sibling---and it grows as the policy improves and more failed steps semantically match a successful reference, so SCPO recovers progressively more credit. The magnitude nonetheless stays small (mean auxiliary reward ${\approx}0.03$, well below the task reward) rather than growing without bound, consistent with monotonic crediting, which credits each reference position at most once (\S\ref{app:bound}). The default reordered allocation credits more than chronological order ($12.0$ vs.\ $9.7$ steps per failed trajectory and $0.034$ vs.\ $0.029$ mean auxiliary reward over steps $20$--$150$), mirroring its $+3.3$ advantage in \S\ref{sec:exp-allocation-order}.

\subsection{Shaper timing}
\label{app:cost-detail}

\begin{table}[H]
\centering
\footnotesize
\setlength{\tabcolsep}{3pt}
\begin{tabular}{@{}lrr@{}}
\toprule
\textbf{Sub-stage} & \textbf{Median (s)} & \textbf{Share} \\
\midrule
Cross-encoder on ref.--failed pairs    & 23.68 & 84.73\% \\
Reference self-similarity scoring      & 4.06  & 14.00\% \\
Tokenizer decode                       & 0.36  & 1.14\% \\
Monotonic credit matching              & 0.004 & 0.02\% \\
Reward accumulation                    & 0.003 & 0.01\% \\
Metric aggregation / tensor conv.      & 0.008 & 0.03\% \\
\bottomrule
\end{tabular}
\caption{SCPO shaper timing decomposition.}
\label{tab:cost-skmp}
\end{table}

As Table~\ref{tab:cost-skmp} shows, nearly all overhead comes from frozen cross-encoder inference; the monotonic credit matcher itself is negligible. The sub-stage medians above sum to ${\approx}28$\,s; the $29.3$\,s figure reported in \S\ref{sec:exp-cost} is the mean wall-clock increase per training step and additionally accounts for inter-process communication with the reranker subprocess.

\end{document}